\documentclass[preprint,12pt]{elsarticle}

\usepackage{amssymb}
\usepackage{amsmath}
\usepackage{ulem}
\usepackage{subfigure}
\usepackage{booktabs}
\newlength\lengtha 
\newlength\lengthb 

\newcommand{\tabincell}[2]{\begin{tabular}{@{}#1@{}}#2\end{tabular}}
\newcommand{\tablefont}{\fontsize{8pt}{\baselineskip}\selectfont}

\usepackage{caption} 
\usepackage{xcolor}

\definecolor{azure(colorwheel)}{rgb}{0.0, 0.5, 1.0}

\journal{}

\begin{document}

\begin{frontmatter}



\title{Anatomical Landmarks Localization for \\3D Foot Point Clouds}


\author[inst]{Sheldon Fung}
\author[inst]{Xuequan Lu}
\author{Mantas Mykolaitis}
\author{Gediminas Kostkevičius}
\author{Domantas Ozerenskis}

\affiliation[inst]{organization={Deakin University},
            city={Geelong},
            country={Australia}}

\begin{abstract}
3D anatomical landmarks play an important role in health research. Their automated prediction/localization thus becomes a vital task. In this paper, we introduce a deformation method for 3D anatomical landmarks prediction. It utilizes a source model with anatomical landmarks which are annotated by clinicians, and deforms this model non-rigidly to match the target model. Two constraints are introduced in the optimization, which are responsible for alignment and smoothness, respectively. Experiments are performed on our dataset and the results demonstrate the robustness of our method, and show that it yields better performance than the state-of-the-art techniques in most cases. 
\end{abstract}


\end{frontmatter}

\section{Introduction}
\label{sec:intro}
As imaging technology becoming more advanced, 3D Volume data is a commonly used format for a variety of fields, especially in the medical and manufacturing industry. Naturally, 3D anatomical landmarks are also becoming significant and prerequisites for those fields. Identifying landmarks is essential for body deformities diagnosis/treatment planning and professional sports equipment (e.g., sneakers) design. However, anatomical landmarks detection via visual assessment of the 3D model requires experienced specialists or clinicians and thus remains a difficult and costly task.

To tackle this problem, a variety of 3D anatomical landmarks detection techniques has been proposed in the previous studies. Some researchers took the advantage of deep learning techniques. For example, Alison et al. \cite{Alison2018} utilized an efficient deep learning method for locating the landmarks in CT scan image data which involves regular pixels. Similar technique is also seen in other papers \cite{Zheng2015,Zhang2017,Bier2018}. However, these methods are vulnerable to irregular 3D point cloud models. Some other methods resort to template model deformation strategy. For example, Fang et al. \cite{Fang1996} introduced a technique utilizing volume morphing for 3D biological point landmark predictions. Whereas such methods often neglect the role of non-rigid deformation which will lead to fewer local errors when matching the source and the target model.

Given the above motivations, we introduce a novel 3D anatomical landmarks prediction approach in this paper. The core mind of our method is to deform a source model with known landmarks (manually annotated) to match the target model, then we can predict the target landmarks according to the transformed known landmarks. We achieve this by resorting to the deformation graph \cite{Sumner2007}. Concretely, inspired by the technique proposed in \cite{Sumner2007}, we first randomly select vertices in a mesh model as nodes and build a deformation graph. Two constraints play the role of the optimization energy function: aligning two models non-rigidly while keeping the smoothness of the template surface. Then, following Lu et al. \cite{Xuequan2018}, we can apply Singular Value Decomposition to obtain the rotation matrices and transformation vectors for the deformation equation. Finally, landmarks predictions can be made via deforming the known landmarks.

Extensive experiments on our private dataset which contains 211 3D foot models with 21 manually annotated anatomical landmarks by professional clinicians to validate our method. We also compare it with ICP (rigid alignment) and two other state-of-the-art non-rigid deformation approaches. Experimental results show that our method is robust and outperforms other methods in most cases, in terms of the root square error metric. Our main contributions are as follows:
\begin{itemize}
    \item We propose a non-rigid deformation method to predict 3D anatomical landmarks on point clouds or meshes.
    \item We conduct experiments on a dataset which contains various shapes of foot models and compare the proposed method with other techniques.
\end{itemize}

The rest of the paper is organized as follows. Section \ref{sec:review} reviews previous research work on 3D landmarks prediction and non-rigid point registration methods. Section \ref{Sec:methodology} illustrates our proposed approach. Detailed experimental results and analysis will be cover in \ref{sec:results}. And section \ref{sec:Conclusion} summarize this work.

\section{Related Work}
\label{sec:review}
In this section, we are first going to review some previous works on 3D landmark prediction. Second, we will cover some state-of-the-art works on non-rigid registration which were developed to align 3D scans of rigid objects which are distorted due to various factors.

\textbf{3D landmark prediction methods.} Most current 3D anatomical landmark prediction methods rely on two categories of data: computed tomography (CT) scan and magnetic resonance imaging (MRI) data. The prediction techniques are various. Some methods are based on deep learning techniques. Zheng et al. \cite{Zheng2015} proposed a two-step approach using a shallow network and a deep network. Similar techniques are also seen in other works. Alison et al. \cite{Alison2018} presented an efficient deep learning method for predicting anatomical landmarks' location in CT scan data, which meets the variability across all landmarks classes. A slightly different approach was introduced by Amir et al. \cite{Alansary2019} which evaluates reinforcement learning (RL) strategies and use RL agent to identify the landmarks by interacting with an environment. With the similar idea, Florin et al. \cite{Ghesu2018} also proposed a deep reinforcement learning (DRL) method to detect anatomical landmarks. More intriguingly, Ebner et al. \cite{Ebner2014} proposed a landmark localization algorithm towards MRI using multiple random regression forests. Creatively, Subburaj et al. \cite{Subburaj2009} presented an anatomical landmark prediction method by segmenting the model surface into different regions based on surface curvature. Some other methods resort to identify the landmarks by deforming the existed annotated model. Fang et al. \cite{Fang1996} introduced a technique using volume morphing for 3D biological point landmark predictions. Baek et al. \cite{Baek2013} developed an anatomically deformable model of the femur based on CT images in order to predict the bone landmarks.

\textbf{Non-rigid registration.} As a widely studied and utilized technique, the solutions for non-rigid registration are diverse. We can roughly categorize them into two groups: optimization-based and statistics-based. For the former one, Sumner et al. \cite{Sumner2007} presented an algorithm that generates natural deformation according to the constraints provided by users. With a similar approach, Li et al. \cite{Li2009} developed a robust framework for reconstruction of complex deforming shapes. It uses a smooth template provides as a coarse approximation of the scanned object which serves as a geometric and topological prior for the reconstruction process. Lu et al.  \cite{Xuequan2018, Xuequan2019} 
introduced an Expectation-Maximization (EM) algorithm for the non-rigid registration of human bodies (point cloud data). Yao et al. \cite{Yao2020} proposed a non-rigid registration algorithm using Welsch's function \cite{Holland1977} for both alignment error and regularization, addressing the drawback of slow convergence due to the high-accuracy solution in other methods \cite{Boyd2011}. From the statistics perspective, Myronenko et al. \cite{Myronenko2010} presented Coherent Point Drift (CPD), a robust probabilistic multidimensional point set registration algorithm for both rigid and non-rigid transformation using Gaussian Mixture Model (GMM). The core mind of this method is to force the GMM to move coherently as a group to the target data points by maximizing the likelihood. As an extension of CPD, Dai et al. \cite{Dai2017} proposed a fully automated pipeline to synthesize shape-and-texture 3D morphable model. Unlike other methods, Wand et al. \cite{Wand2009} introduced a novel topology-aware adaptive subspace deformation technique which no longer requires model template.

\section{Method}
\label{Sec:methodology}

\begin{figure}[thbp]
\centering
\includegraphics[width=1\linewidth]{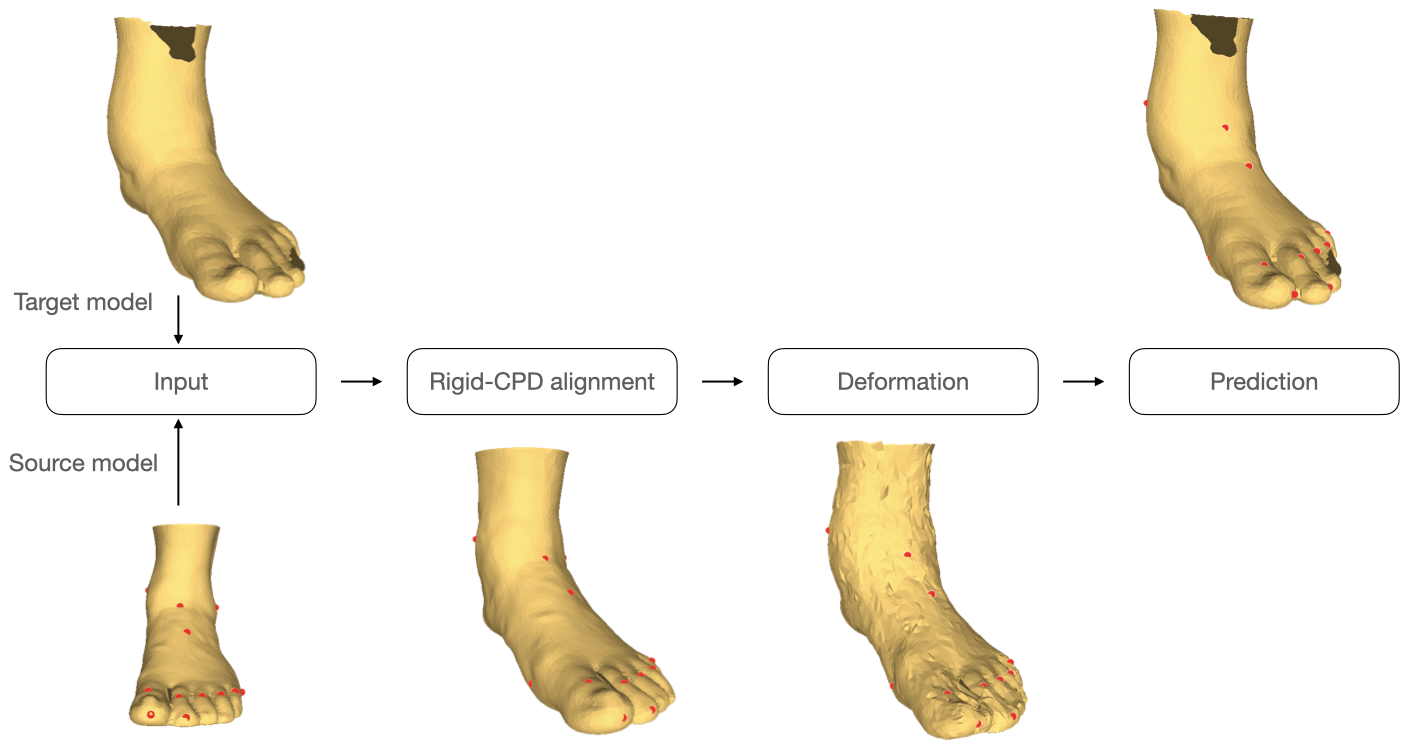}
\caption{Overview of our proposed method. Note that the red points represent the landmarks of the corresponding foot model.} \label{fig:overview} 
\end{figure}

In this section, we will introduce our approach in detail. As illustrated in Figure \ref{fig:overview}, we first take both the annotated source model and the target model as input and align the former to the latter with rigid-CPD. Then we deform the source model with a deformation graph to match the target model. Finally, we can predict the landmarks on the target model based on the known landmarks on the source model.

\subsection{Deformation Formulation}
The goal of our method of is to realize landmark prediction on the surface of 3D models (e.g. meshes or point clouds) using an existing 3D model with known ground-truth landmarks via non-rigid registration. Let $V_t=\{v_{t1}, v_{t1}, \cdots, v_{tn} \in \mathbb{R}^3\}$ be the target model and
let $V_{sr}=\{V_s, R_l\}$ be the point set consisting of source model $V_s=\{v_1, v_2, \cdots, v_n \in \mathbb{R}^3\}$ with manually annotated landmarks $R_l=\{r_1, r_2,\cdots,r_n \in \mathbb{R}^3\}$. Note that here $R_l$ is not necessary to be a subset of $V_s$. To ensure these two models are well aligned to avoid misleading results of closest point search in the following steps, we initialize $V_{sr}$ and $V_t$ with the global rigid point set registration in CPD \cite{Myronenko2010}. To deform the source model $V_s$ to match the target model $V_t$, we adopt the deformation graph following Sumner et al. \cite{Sumner2007}. Concretely, we randomly sample the source model $V_s$ to construct the nodes $N=\{n_1,n_2,\cdots,n_j\in \mathbb{R}^3\}$. For each node $n_j \in N$, the affine transformation is specified by a rotation matrix $\mathbf{R} \in \mathbb{R}^{3\times 3}$ and a translation vector $\mathbf{T} \in \mathbb{R}^{3\times 1}$. Each vertex $v_n \in V_s$ is induced by the combined influence of the neighbouring $k$ nearest nodes, as shown in Eq. \eqref{eq:deformationAl1}.
\begin{equation}
    \label{eq:deformationAl1}
    v_n^{\prime}=\sum_{n_j \in \Bbbk(v_n)}^k \omega_{j}(v_n)[\mathbf{R}_j(v_n-n_j)+n_j+\mathbf{T}_j],
\end{equation}
where $v_n^{\prime}$ is the transformed position of $v_n$, and $\Bbbk(v_n)$ denotes the $k$ nearest nodes of $v_n$. Notice that there are different approaches for searching the close nodes, such as Euclidean distance and Breadth-first search (BFS). In our experiments, we adopt the former strategy.  

The weight between $n_j$ and $v_n$ is defined as follows: 
\begin{equation}
    \omega_{j}(v_n)=\mathbb{D}_{nj}^{-1}/d
\end{equation}
where $\mathbb{D}_{nj}$ and $d$ are defined in Eq. \eqref{eq:weights_D} and \eqref{eq:weights_d}, respectively.
\begin{equation}
\label{eq:weights_D}
    d=\sum_{n_j\in \Bbbk(v_n)}\frac{1}{|v_n-n_j|}
\end{equation}
\begin{equation}
\label{eq:weights_d}
    \mathbb{D}_{n_j}= \begin{cases} 
    |v_n-n_j|, &(v_n \neq n_j)\\
    \sum_{n_j\in \Bbbk(v_n)}|v_n-n_j|, &(v_n = n_j)
    \end{cases}
\end{equation}

\subsection{Optimization}

We now introduce the optimization strategy for the involved parameters $\textbf{R}_j$ and $\textbf{T}_j$ in the deformation equation. We adopt two constraints for the energy function: data term and smooth term. 

\textbf{Data Term.} To non-rigidly align the template model to the target model, we design a non-rigid Iterative Closest Point (ICP)  method. For each vertex $v_n \in V_s$, we look for the closest vertex $c_k \in V_t$. And the alignment term minimizes the deviation between the corresponding transformed vertex $v_n^{\prime}$ and the closest vertex $c_k$. Following previous works \cite{Sumner2007, Besl1992}, we solve it as a $\ell_2$ minimization problem with the following objective equation:
\begin{equation}
\label{EqFitwithym}
    E_{align} = \sum_{(v_n,c_k) \in C} \|v_n^{\prime}-c_k\|_2^2
\end{equation}
where $C$ is the correspondences (closest pairs) between $V_s$ and $V_t$. Note that using  $\ell_2-norm$ for minimizing the point-wise distance brings about noisy surface on the transformed model due to the noisy data or incomplete structure induced by the camera scans. However, our goal is to force the template model to non-rigidly approach the target model, via which we make the landmark predictions. Therefore, the smoothness of the model surface is not considered to be a top priority in practice. 

\textbf{Smooth term.} Each node serves as a localized leader point, influencing all other vertices in the local sense. Therefore, nodes should be resistant to the overlapping influence of each other. Specifically, consider a specific node $n_j\in N$ and its one nearby node $n_k$. The influence $I_j$ and $I_k$ which are respectively induced by $n_j$ itself and $n_k$ can be computed as follows:
\begin{equation}
    I_j=\mathbf{R}_k(n_j-n_j)+n_j+\mathbf{T}_k=(n_j+T_k)
\end{equation}
\begin{equation}
    I_k=\mathbf{R}_k(n_j-n_k)+n_k+\mathbf{T}_k
\end{equation}
the transformation of $n_k$ should be the weighted sum of each influence, which is computed as:
\begin{equation}
    n_j^\prime=\omega_j(n_k)I_k+\omega_j(n_k)I_j
\end{equation}

We hope that the transformation of $n_j$ to be resistant to the influence of $n_k$ without eliminating its affect by forcing $I_k$ to agree with $I_j$. Consequently, $n_j^\prime$ can thus be theoretically equivalent to $W(n_k+T_k)$, where $W=w_k(n_k)+w_j(n_k)$, which is derived from Eq. \eqref{eq:deformationAl1}. This is achieved by minimizing the $\ell_2$ distance for $I_k$ of node and $I_j$. In practice, the regularization term should account for each node in the deformation graph with all their neighboring nodes:
\begin{equation}
\label{EqSmooth}
    E_{smooth}=\sum_{n_j}\sum_{n_k\in \Bbbk(n_j)}\|I_j-I_k\|_2^2,
\end{equation}

Finally, we define the total energy function $E_{total}$ by combining the alignment function and the regularization function with a coefficient $\alpha$. Note that $R_j$ should be restricted to be in SO(3) with the orthogonal constraint $R_j^TR_j=I, \forall j$.

\begin{equation}
\label{energy_total}
    E_{total}=E_{smooth}+\alpha E_{align}
\end{equation}

\subsection{Minimization}
To efficiently solve the rotation matrix $R$ as well as the translation vector $T$ for each node transformation, we minimize $E_{total}$ following Lu et al. \cite{Lu2018} and take the advantage of the following scheme: with the fixing remaining nodes in the deformation graph, we can update an individual node instead at each time, through which not only the computational complexity can be enormously reduced, but also the non-positive growing trend of the energy function can be guaranteed.

We can obtain the following equation by taking the derivative of Eq. \eqref{energy_total} with respect to the translation vector $T_j$ of a specific node $j$ and equate it to zero:
\begin{equation}
\label{eq:derivative} 
    \textbf{T}_j=\mu_v-\textbf{R}_j\mu_y,
\end{equation}
where $\mu_v, \mu_y \in \mathbb{R}^{3\times 1}$ represent the rest parts of the equation which can be calculated. Next, we substitute \eqref{eq:derivative} to \eqref{energy_total}, then organize it to obtain: $E=-tr(\textbf{H}\textbf{R}_j)+Z_j, \textbf{H}\in R^{3\times 3}$, where $tr(\cdot)$ and $Z_j$ are the trace operator and a scalar, respectively. Following the work  \cite{Myronenko2009}, we resort to Lemma 1 for achieving the closed-form solution of $R_j$ which will maximize $-E$ (equivalent to minimizing $E$):
\begin{equation}
    \mathbf{R}_{\hat j}=\mathbf{U}_{\hat j}\mathbf{C}_{\hat j}\mathbf{V}_{\hat j}^T,
    \label{eq:computeR1}
\end{equation}
where $\mathbf{C}_{\hat j}=\mathit{diag}(1, 1, \mathit{det}(\mathbf{U}_{\hat j}\mathbf{V}_{\hat j}^T))$. We apply Singular Value Decomposition (SVD) on $H^T$ to obtain $\mathbf{U}_{\hat j}\mathbf{S}_{\hat j}\mathbf{V}_{\hat j}^T=\mathit{svd}(\mathbf{H}_{\hat j}^T)$.  Finally, the translation vector $T_j$ of node $j$ can be easily calculated from Eq. \eqref{eq:derivative}.

\section{Experimental Results}
\label{sec:results}
In this section, we will first illustrate the details of the implementation used in our experiments. Then we will introduce the test data and error metric we use in our experiments. In the end, we will show our experimental results and compare them with the results using state-of-the-art methods.

\subsection{Implementation Details}
The implementation of our method is written in C++ with the functionality provided by libigl \cite{libigl2018} and Eigen \cite{eigen2018}. All the experiments are conducted on a computer equipped with a Quad-Core Intel Core i5 CPU (2.3 GHz) and 8GB memory.  
For the parameters in the experiments, we set the node number to be 500 and each vertex is set to be influenced by 10 nearest nodes. The $\alpha$ in the objective function (see Eq. \eqref{energy_total}) is set to 2000.

\subsection{Database and Data Preprocessing}
\label{Experimental:database}
\begin{figure}[hbt!]
\centering
\begin{minipage}[b]{0.24\linewidth}
\subfigure[]{\label{fig:footModel:a}\includegraphics[width=1\linewidth]{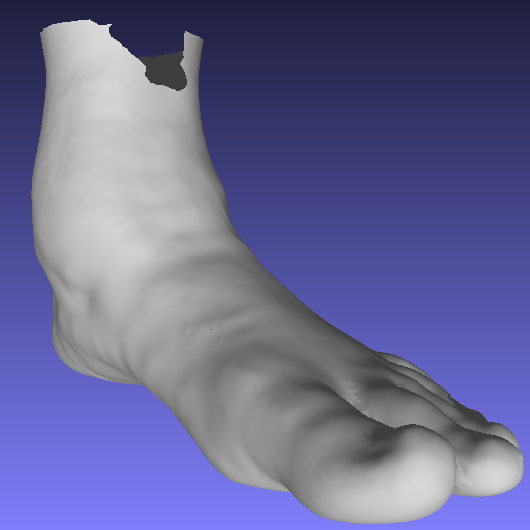}}
\end{minipage}
\begin{minipage}[b]{0.24\linewidth}
\subfigure[]{\label{fig:footModel:b}\includegraphics[width=1\linewidth]{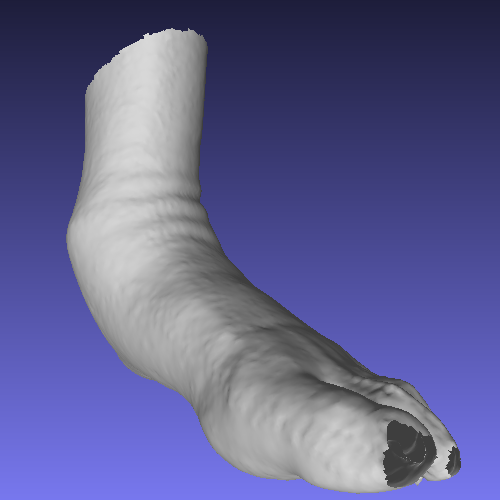}}
\end{minipage}
\begin{minipage}[b]{0.24\linewidth}
\subfigure[]{\label{fig:footModel:c}\includegraphics[width=1\linewidth]{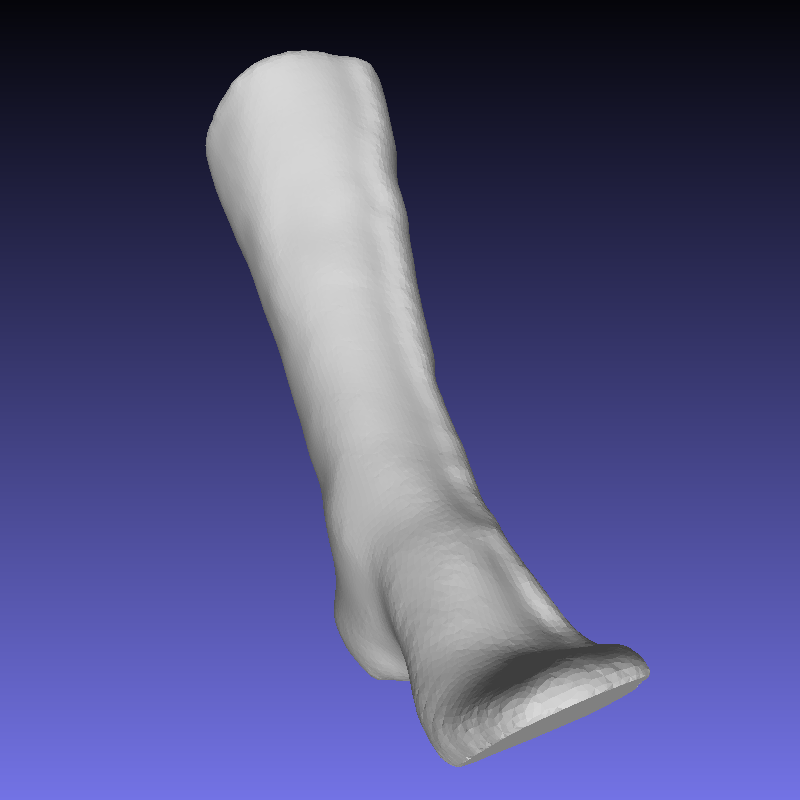}}
\end{minipage}
\begin{minipage}[b]{0.24\linewidth}
\subfigure[]{\label{fig:footModel:d}\includegraphics[width=1\linewidth]{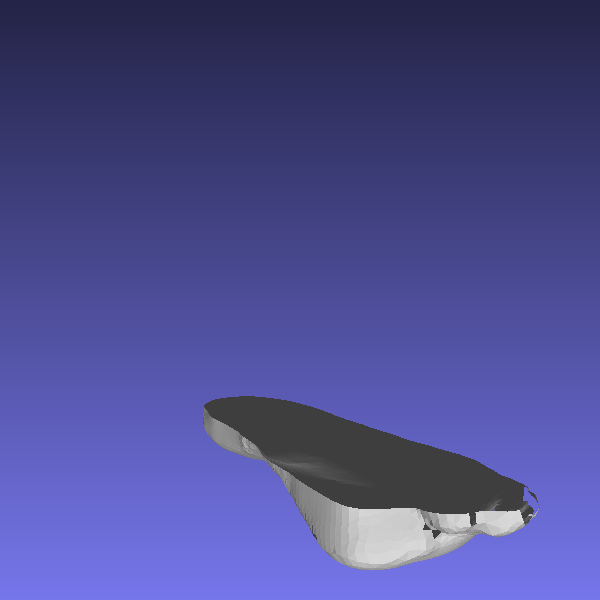}}
\end{minipage}
\caption{Foot mesh model examples with various shapes from the database.}
\label{fig:footModel}
\end{figure}

The database contains 1,511 real-world 3D-scan foot mesh models with 21 ground-truth landmarks which are marked by medical specialists. Figure \ref{fig:footModel} displays the foot models sampled from the database. Notice the huge diversity of the foot shapes: Figure \ref{fig:footModel:b} is a foot model wearing socks during scanning; the foot model of Figure \ref{fig:footModel:c} consists of an additional leg area which is not the region of interest; due to human error during the scanning, the model in Figure \ref{fig:footModel:d} only contains the bottom part of the foot. Fortunately, cases like Figure \ref{fig:footModel:c} and \ref{fig:footModel:d} take up only a negligible proportion in the database.

We randomly select approximately 20\% of the database forming a test dataset for our experiments, which contains 211 foot models.  The original foot model contains approximately 50,000 vertices including duplicated elements. For efficiency reasons, we decimate the meshes with Meshlab \cite{meshlab2008} to 5,000 vertices per model. 

\begin{figure}[hbt!]
\centering
\begin{minipage}[b]{0.24\linewidth}
\subfigure[]{\label{fig:footModel_lm:a}\includegraphics[width=1\linewidth]{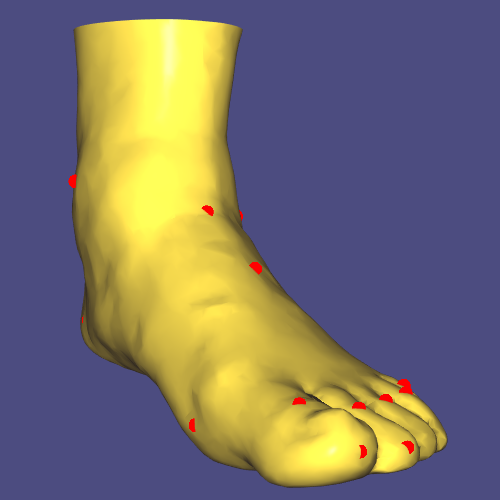}}
\end{minipage}
\begin{minipage}[b]{0.24\linewidth}
\subfigure[]{\label{fig:footModel_lm:b}\includegraphics[width=1\linewidth]{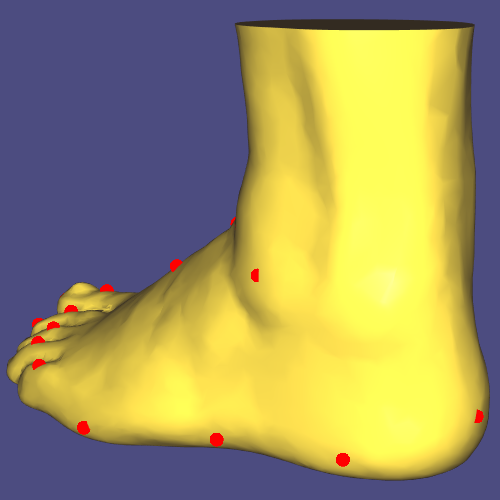}}
\end{minipage}
\begin{minipage}[b]{0.24\linewidth}
\subfigure[]{\label{fig:footModel_lm:c}\includegraphics[width=1\linewidth]{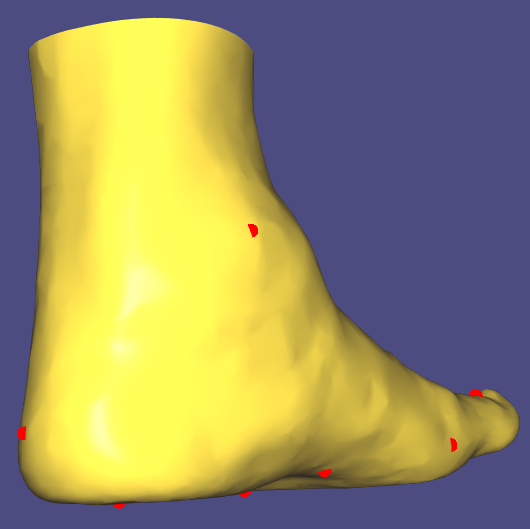}}
\end{minipage}
\begin{minipage}[b]{0.24\linewidth}
\subfigure[]{\label{fig:footModel_lm:d}\includegraphics[width=1\linewidth]{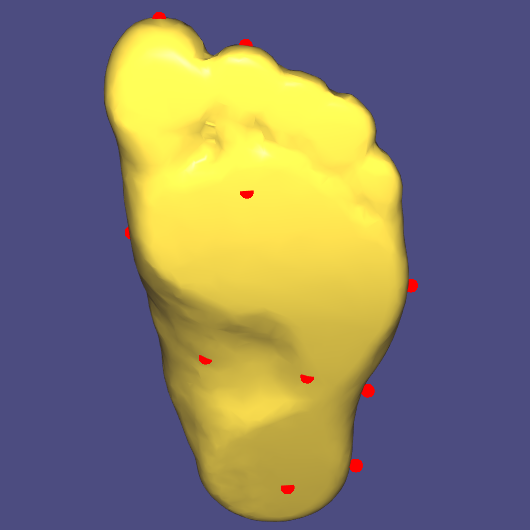}}
\end{minipage}
\caption{four different views of a foot model after data preprocessing along with the anatomical landmarks annotated by medical specialists.}
\label{fig:footModel_lm}
\end{figure}

Figure \ref{fig:footModel_lm} displays different views of a single foot model after data preprocessing (contains 5,000 vertices) with manually annotated anatomical landmarks. Notice that, this specific foot model which originally contains 34,395 vertices does not belong to the test dataset and is chosen to be the source model throughout the experiments, as its shape is visually more universal.

\subsection{Error Metric}
\label{subsec:errMetric}
To measure the performance of the landmark predictions, we use the average of the square norm between the target landmarks and the predicted landmarks, shown as follows:
\begin{equation}
    Err_{avg} = \frac{1}{n}\sum_{i}^n\|Lp_{i}-Lt_{i}\|_2,
    \label{estimate2}
\end{equation}
where $Lp_{i}$ and $Lt_{i}$ are the $i$-th predicted landmark and the target landmark, respectively.

\subsection{Comparison With State-of-the-art Methods}
\textbf{Source model.} The source foot model (see Section \ref{Experimental:database}) is processed to different resolutions with Meshlab \cite{meshlab2008}. To be specific, four resolutions containing different amount of vertices (2k, 5k, 10k, original) are selected to perform the experiments.

\textbf{Other methods.} In the experiments, three high-performing point set registration methods are chosen to be the competitors to our method. Note that experiments with some source model resolutions are skipped while using the original source codes of non-rigid CPD \cite{Myronenko2010} and Fast-RNRR \cite{Yao2020}, in order to avoid the occurrence of certain issues of their codes (i.e. prompting errors).

\begin{table}[hbt!]\tablefont
    \centering
    \caption{Comparisons. 
    }\label{table:comparisons}
    \begin{tabular}{@{} l
                @{\hspace*{\lengthb}}c
                @{\hspace*{\lengthb}}c
                @{\hspace*{\lengthb}}c
                @{\hspace*{\lengthb}}c
                @{\hspace*{\lengthb}}c
                @{\hspace*{\lengthb}}c
                @{\hspace*{\lengthb}}c
                @{\hspace*{\lengthb}}c}
    \toprule
    Resolution & \multicolumn{2}{c}{
    2k
    } & \multicolumn{2}{c}{
    5k
    } & \multicolumn{2}{c}{
    10k
    } & \multicolumn{2}{c}{
    original (34k)
    }\\ 
    \midrule
    Method & \tabincell{c}{Avg.
    } & \tabincell{c}{Mid.
    } & \tabincell{c}{Avg.
    } & \tabincell{c}{Mid.
    } & \tabincell{c}{Avg.
    } & \tabincell{c}{Mid.
    } & \tabincell{c}{Avg.
    } & \tabincell{c}{Mid.
    }\\
    \midrule
    ICP & 17.81 & 15.88 & 17.76 & 15.90 & 17.76 & 15.83 & 17.76 & 15.88
    \\
    non-rigid CPD & 15.72 & 15.17 & 15.73 & 15.21 & - & - & - & -
    \\
    Fast-RNRR & - & - & - & - & - & - & 17.92 & 15.11
    \\
    \midrule
    our method & 15.03 & 14.39 & 14.90 & 14.15 & 15.05 & 14.22 & 14.73 & 13.98
    \\
    \bottomrule
    \end{tabular}
\end{table} 

In Table \ref{table:comparisons}, we report the average (Avg.) and the median (Mid.) of the error results using the metric mentioned in \ref{subsec:errMetric}. We can observe that our method overall outperforms other methods by a noticeable margin. Our method reaches the best performance when using the original source data, which outweighs the ICP and Fast-RNRR by 3.03 and 3.19 for average errors, respectively. Our method also outperforms other method when using different resolutions. The errors for ICP are 2.78, 2.86 and 2.71 higher than our method when tested on 2k, 5k and 10k, respectively. Despite that Fast-RNRR saw a strong performance for 2k and 5k resolution, reaching 15.72 and 15.17, respectively. The errors of our method are still 0.69 and 0.83 lower, respectively. Another phenomenon we observed from the results of different resolutions is that the errors are negatively correlated with the resolutions of the target models.

\subsection{Case Study}

\begin{figure}[hbt!]
\centering
\begin{minipage}[b]{0.19\linewidth}
\subfigure[]{\label{fig:footModel_lm:a}\includegraphics[width=1\linewidth]{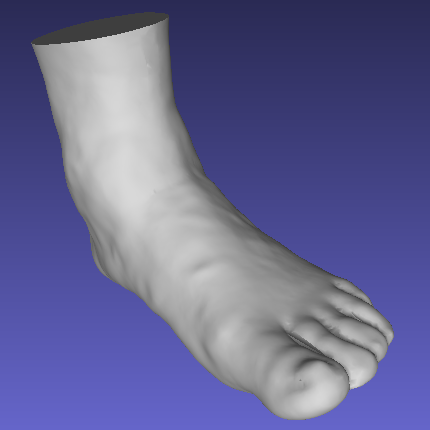}}
\end{minipage}
\begin{minipage}[b]{0.19\linewidth}
\subfigure[]{\label{fig:footModel_lm:c}\includegraphics[width=1\linewidth]{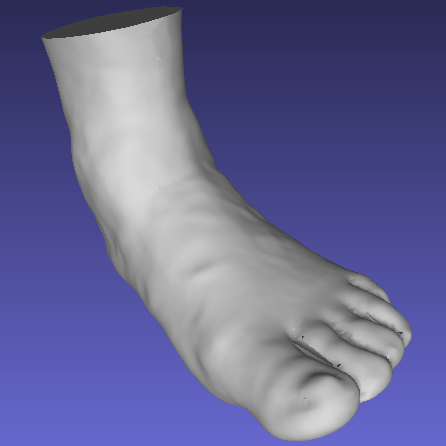}}
\end{minipage}
\begin{minipage}[b]{0.19\linewidth}
\subfigure[]{\label{fig:footModel_lm:b}\includegraphics[width=1\linewidth]{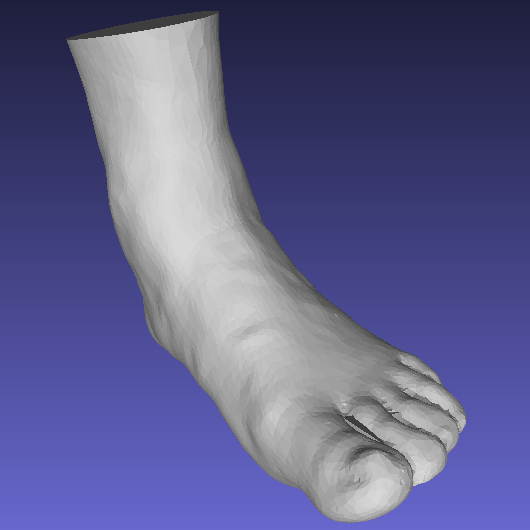}}
\end{minipage}
\begin{minipage}[b]{0.19\linewidth}
\subfigure[]{\label{fig:footModel_lm:b}\includegraphics[width=1\linewidth]{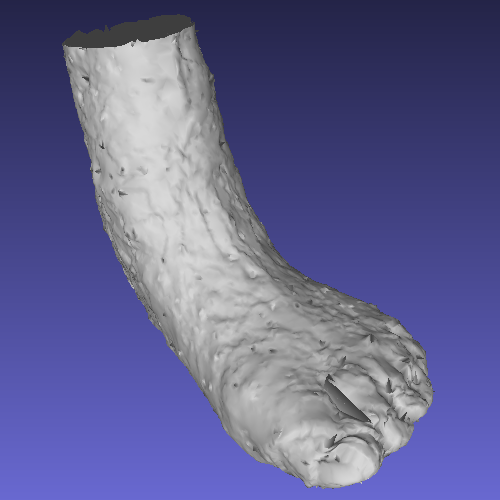}}
\end{minipage}
\begin{minipage}[b]{0.19\linewidth}
\subfigure[]{\label{fig:footModel_lm:d}\includegraphics[width=1\linewidth]{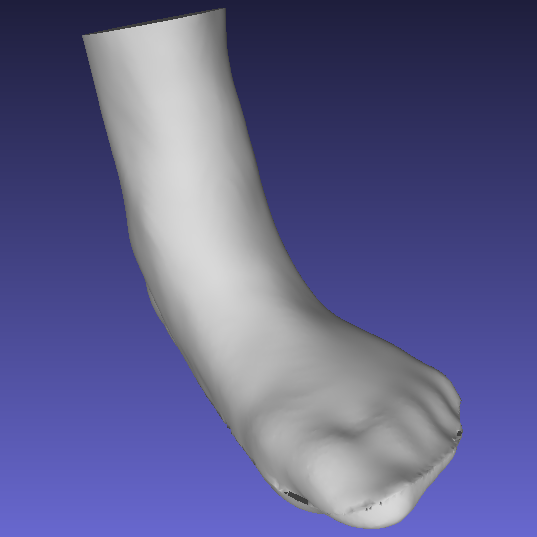}}
\end{minipage}

\begin{minipage}[b]{0.19\linewidth}
\subfigure[]{\label{fig:footModel_lm:a}\includegraphics[width=1\linewidth]{figures/cases/ori.png}}
\end{minipage}
\begin{minipage}[b]{0.19\linewidth}
\subfigure[]{\label{fig:footModel_lm:c}\includegraphics[width=1\linewidth]{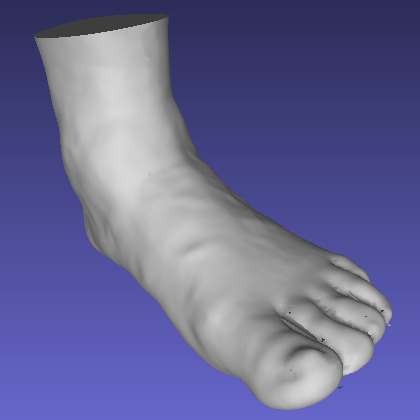}}
\end{minipage}
\begin{minipage}[b]{0.19\linewidth}
\subfigure[]{\label{fig:footModel_lm:b}\includegraphics[width=1\linewidth]{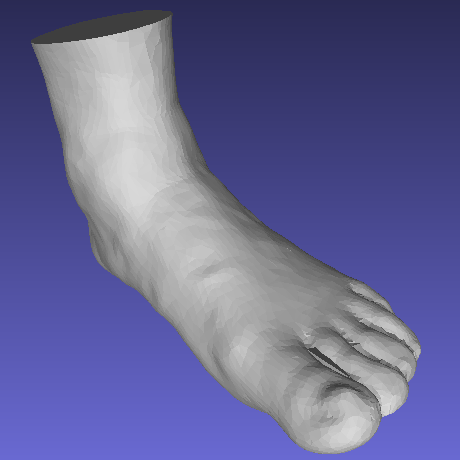}}
\end{minipage}
\begin{minipage}[b]{0.19\linewidth}
\subfigure[]{\label{fig:footModel_lm:b}\includegraphics[width=1\linewidth]{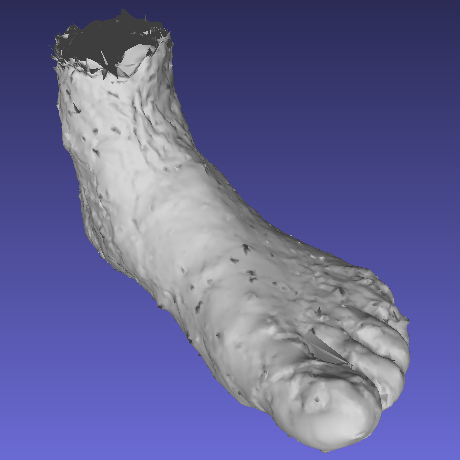}}
\end{minipage}
\begin{minipage}[b]{0.19\linewidth}
\subfigure[]{\label{fig:footModel_lm:d}\includegraphics[width=1\linewidth]{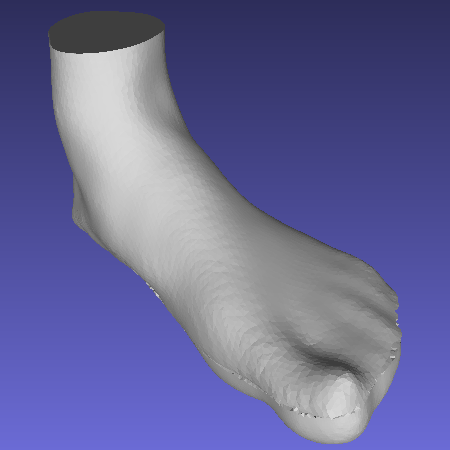}}
\end{minipage}

\begin{minipage}[b]{0.19\linewidth}
\subfigure[]{\label{fig:footModel_lm:a}\includegraphics[width=1\linewidth]{figures/cases/ori.png}}
\end{minipage}
\begin{minipage}[b]{0.19\linewidth}
\subfigure[]{\label{fig:footModel_lm:c}\includegraphics[width=1\linewidth]{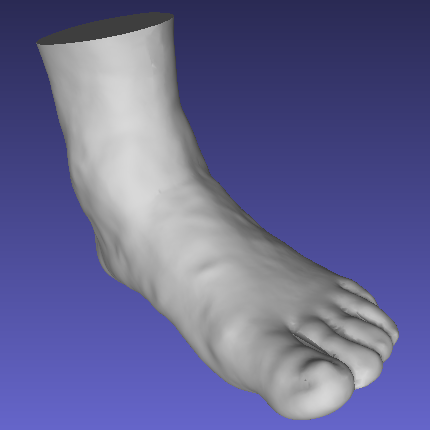}}
\end{minipage}
\begin{minipage}[b]{0.19\linewidth}
\subfigure[]{\label{fig:footModel_lm:b}\includegraphics[width=1\linewidth]{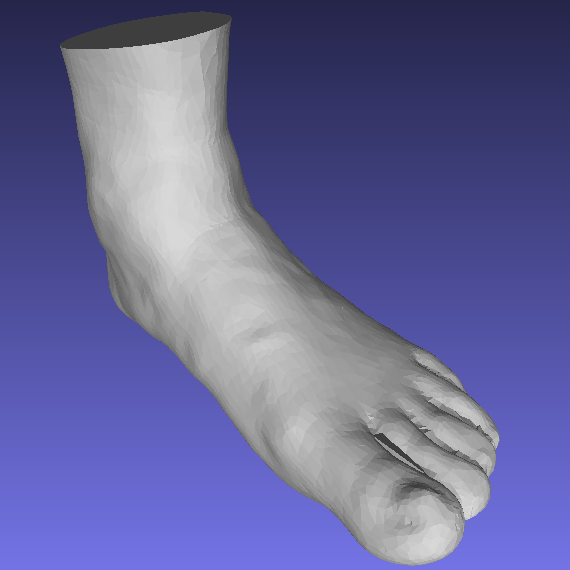}}
\end{minipage}
\begin{minipage}[b]{0.19\linewidth}
\subfigure[]{\label{fig:footModel_lm:b}\includegraphics[width=1\linewidth]{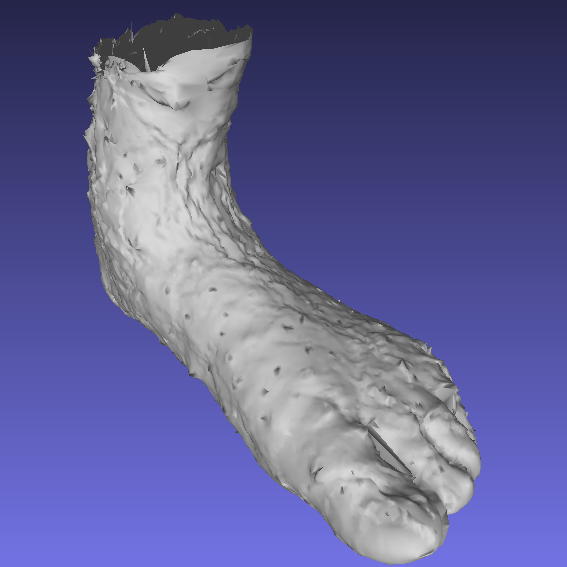}}
\end{minipage}
\begin{minipage}[b]{0.19\linewidth}
\subfigure[]{\label{fig:footModel_lm:d}\includegraphics[width=1\linewidth]{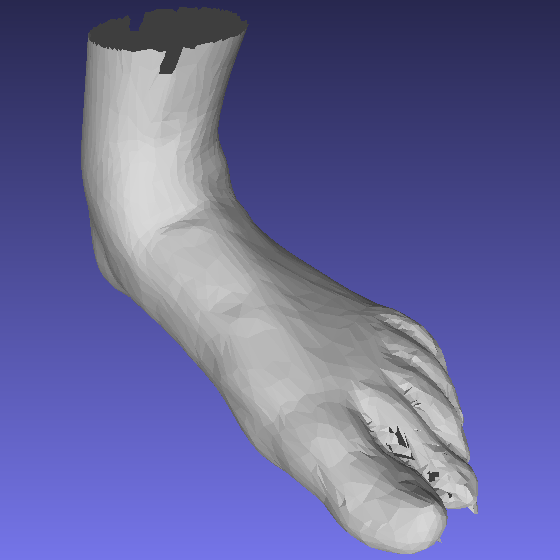}}
\end{minipage}
\caption{(a), (f) and (k) are the source models. (e), (j) and (o) are the target models with ID 43, 711 and 716, respectively, which are randomly selected from the test set. (b), (g) and (l) are the models synthesize with Fast-RNRR. (c), (h), and (m) are the models synthesize with non-rigid CPD. (d), (i), and (n) are the models synthesize with our method.}
\label{fig:footModelsComp}
\end{figure}

We randomly select three target models from the test set and compare the synthesized mesh models using Fast-RNRR, CPD, and our method in Figure \ref{fig:footModelsComp}. CPD and Fast-RNRR have a common characteristic that they tend to preserve the original shape of the source model, which will probably lead to larger errors. We can observe that the model synthesized with our method can better fit the target model. Concretely, regarding the foot model with ID 43, we can notice the narrow footbridge and the small heel are reflected on the model synthesized with our method. In the other two cases, we can also observe how our method outperforms the other two methods in such detailed aspects (e.g., the narrow ankle area for the model with ID 711 and the forward-bending ankle for the model with ID 716). One minor drawback of our method is the high surface roughness of the generated models. Nevertheless, the goal of our method focuses on accurate landmarks prediction, therefore this weakness is negligible. 

\begin{table}[thbp]\tablefont
    \begin{center}
    \caption{Corresponding errors of figure \ref{fig:footModelsComp}}
    \label{table:cErrors}
    \begin{tabular}{@{} l
                @{\hspace*{\lengtha}}c
                @{\hspace*{\lengtha}}c
                @{\hspace*{\lengtha}}c
                @{\hspace*{\lengtha}}c}
    \toprule
    model ID & \tabincell{l}{
    43
    } & \tabincell{l}{
    711
    } & \tabincell{l}{
    716
    }\\ 
    \midrule
    CPD & 10.95 & 9.35 & \textbf{10.47}
    \\
    Fast-RNRR & 9.25 & 11.71 & 31.28
    \\
    ours & \textbf{8.90} & \textbf{8.97} & 10.82
    \\
    \bottomrule
    \end{tabular}
    \end{center}
\end{table}

Statistically, the effectiveness of our method is also demonstrated by the errors displayed in Table \ref{table:cErrors}. The errors show that our method achieves top results for models with ID 43 and 711, reaching 8.90 and 8.97, respectively. Even though for the model with ID 716 where CPD outperforms our method by 0.35, our approach still outperforms Fast-RNRR by a large margin (20.81), which illustrates the robustness of our method. Further, we also notice that the similarity between the source model and the target model is negatively correlated with the resulting error. Therefore, one limitation is, to minimize the prediction errors, we can choose the template model with a more similar shape to the target model to predict the landmarks.

\section{Discussion \& Conclusion}
\label{sec:Conclusion}
We have presented a robust 3D anatomical landmark detection method for 3D foot models. It realizes the prediction of anatomical landmarks by deforming a given source model with annotated landmarks to a target model. After non-rigid deformation, the updated landmarks of the source model are naturally viewed as the landmarks of the target model. Experiments demonstrate that compared to the state-of-the-art point registration techniques, our method can predict landmarks on target models with fewer errors. As future work, it will be interesting to extend our framework to other landmark prediction tasks such as the human face and human body.


\begin{thebibliography}{00}
\bibitem{Zhang2017}
Zhang, Jun, Mingxia Liu, and Dinggang Shen. "Detecting anatomical landmarks from limited medical imaging data using two-stage task-oriented deep neural networks." IEEE Transactions on Image Processing 26.10 (2017): 4753-4764.

\bibitem{Bier2018}
Bier, Bastian, et al. "X-ray-transform invariant anatomical landmark detection for pelvic trauma surgery." International Conference on Medical Image Computing and Computer-Assisted Intervention. Springer, Cham, 2018.

\bibitem{Ebner2014}
Ebner, Thomas et al. “Towards automatic bone age estimation from MRI: localization of 3D anatomical landmarks.” Medical image computing and computer-assisted intervention : MICCAI ... International Conference on Medical Image Computing and Computer-Assisted Intervention vol. 17,Pt 2 (2014): 421-8. doi:10.1007/978-3-319-10470-6\_53

\bibitem{Baek2013}
Baek, Seung-Yeob, et al. "Automated bone landmarks prediction on the femur using anatomical deformation technique." Computer-aided design 45.2 (2013): 505-510.

\bibitem{Sumner2007}
Sumner, Robert W., Johannes Schmid, and Mark Pauly. "Embedded deformation for shape manipulation." ACM SIGGRAPH 2007 papers. 2007. 80-es.

\bibitem{Fang1996}
Fang, Shiaofen, Raghu Raghavan, and Joan T. Richtsmeier. "Volume morphing methods for landmark-based 3D image deformation." Medical Imaging 1996: Image Processing. Vol. 2710. International Society for Optics and Photonics, 1996.

\bibitem{Subburaj2009}
Subburaj, K., Bhallamudi Ravi, and Manish Agarwal. "Automated identification of anatomical landmarks on 3D bone models reconstructed from CT scan images." Computerized Medical Imaging and Graphics 33.5 (2009): 359-368.

\bibitem{Alison2018}
O'Neil, Alison Q., et al. "Attaining human-level performance with atlas location autocontext for anatomical landmark detection in 3D CT data." Proceedings of the European Conference on Computer Vision (ECCV) Workshops. 2018.

\bibitem{Alansary2019}
Alansary, Amir, et al. "Evaluating reinforcement learning agents for anatomical landmark detection." Medical image analysis 53 (2019): 156-164.

\bibitem{Zheng2015}
Zheng, Yefeng, et al. "3D deep learning for efficient and robust landmark detection in volumetric data." International conference on medical image computing and computer-assisted intervention. Springer, Cham, 2015.

\bibitem{Ghesu2018}
Ghesu, Florin C., et al. "Robust multi-scale anatomical landmark detection in incomplete 3D-CT data." International Conference on Medical Image Computing and Computer-Assisted Intervention. Springer, Cham, 2017.

\bibitem{Li2009}
Li, Hao, et al. "Robust single-view geometry and motion reconstruction." ACM Transactions on Graphics (ToG) 28.5 (2009): 1-10.

\bibitem{Yao2020}
Yao, Yuxin, et al. "Quasi-Newton solver for robust non-rigid registration." Proceedings of the IEEE/CVF conference on computer vision and pattern recognition. 2020.

\bibitem{Holland1977}
Holland, Paul W., and Roy E. Welsch. "Robust regression using iteratively reweighted least-squares." Communications in Statistics-theory and Methods 6.9 (1977): 813-827.

\bibitem{Boyd2011}
Boyd, Stephen, Neal Parikh, and Eric Chu. Distributed optimization and statistical learning via the alternating direction method of multipliers. Now Publishers Inc, 2011.

\bibitem{Brown2007}
Brown, Benedict J., and Szymon Rusinkiewicz. "Global non-rigid alignment of 3-D scans." ACM SIGGRAPH 2007 papers. 2007. 21-es.

\bibitem{Myronenko2010}
Myronenko, Andriy, and Xubo Song. "Point set registration: Coherent point drift." IEEE transactions on pattern analysis and machine intelligence 32.12 (2010): 2262-2275.

\bibitem{Dai2017}
Dai, Hang, et al. "A 3d morphable model of craniofacial shape and texture variation." Proceedings of the IEEE International Conference on Computer Vision. 2017.

\bibitem{Wand2009}
Wand, Michael, et al. "Efficient reconstruction of nonrigid shape and motion from real-time 3d scanner data." ACM Transactions on Graphics (TOG) 28.2 (2009): 1-15.

\bibitem{Xuequan2018}
Lu, Xuequan, et al. "Unsupervised articulated skeleton extraction from point set sequences captured by a single depth camera." Proceedings of the AAAI Conference on Artificial Intelligence. Vol. 32. No. 1. 2018.

\bibitem{Xuequan2019}
Lu, Xuequan, et al. "3D articulated skeleton extraction using a single consumer-grade depth camera." Computer Vision and Image Understanding 188 (2019): 102792.

\bibitem{Besl1992}
Besl, Paul J., and Neil D. McKay. "Method for registration of 3-D shapes." Sensor fusion IV: control paradigms and data structures. Vol. 1611. International Society for Optics and Photonics, 1992.

\bibitem{Lu2018}
Lu, Xuequan, et al. "Unsupervised articulated skeleton extraction from point set sequences captured by a single depth camera." Proceedings of the AAAI Conference on Artificial Intelligence. Vol. 32. No. 1. 2018.

\bibitem{Myronenko2009}
Myronenko, Andriy, and Xubo Song. "On the closed-form solution of the rotation matrix arising in computer vision problems." arXiv preprint arXiv:0904.1613 (2009).

\bibitem{libigl2018}
Alec Jacobson, et al. " libigl: A simple C++ geometry processing library." (2018).

\bibitem{eigen2018}
Ga\"el Guennebaud, Beno\^it Jacob, \& others. (2010). Eigen v3. .

\bibitem{meshlab2008}
Cignoni, Paolo et al. “MeshLab: an Open-Source Mesh Processing Tool.” Eurographics Italian Chapter Conference (2008).

\end{thebibliography}
\end{document}